\documentclass[nonacm]{acmart}
\usepackage{tikz}
\usepackage{pgfplots}

\begin{document}
\bibliographystyle{unsrt}
\title[]{Smart Active Sampling to enhance Quality Assurance Efficiency}
\author{Clemens Heistracher}
\authornote{Both authors contributed equally to this research.}

\affiliation{%
  \institution{AIT - Austrian Institute of Technology}
  \country{Austria}
}
\email{clemens.heistracher[at]ait.ac.at}

\author{Stefan Stricker}
\authornotemark[1]
\affiliation{%
  \institution{STIWA Holding GmbH }
  \country{Austria}
}
\email{stefan.stricker[at]gmail.com}

\author{Pedro Casas}

\affiliation{%
  \institution{AIT - Austrian Institute of Technology}
  \country{Austria}
}
\email{pedro.casas[at]ait.ac.at}

\author{Daniel Schall}
\author{Jana Kemnitz}

\affiliation{%
  \institution{Siemens AG Austria}
  \country{Austria}
}
\email{jana.kemnitz[at]siemens.com}
\email{daniel.schall[at]siemens.com}

\renewcommand{\shortauthors}{Heistracher and Stricker, et al.}

\begin{abstract}
    We propose a new sampling strategy, called smart active sapling, for quality inspections outside the production line. Based on the principles of active learning a machine learning model decides which samples are sent to quality inspection. On the one hand, this minimizes the production of scrap parts due to earlier detection of quality violations. On the other hand, quality inspection costs are reduced for smooth operation. 
\end{abstract}






\maketitle

\section{Introduction}
The quality of products is crucial for a company's success and customers expect a consistently high level. Therefore, quality control plays an important role. However, quality control is also a cost factor that can not be neglected \cite{QIN2015907}. This is especially true for those industries requiring inspections outside the production line, such as disruptive or non-disruptive testing \cite*{tuominen2012cost}. Traditional sampling strategies are random sampling, where a fixed number of samples is taken randomly in a fixed interval or periodic sampling, where samples are taken at a fixed interval. However, both sampling strategies entail some disadvantages. A small sampling interval increases the chance of detecting defective parts but also increases sampling costs, especially using destructive testing. Contrary, a large sampling period decreases costs, but also decreases the chance of detecting those defective parts. Further, this may result in prolonged production of defective parts due to the temporal delay of inspection results. These traditional sampling strategies do not include any information about the individual pieces or the production process itself. For example, changes in production parameters or production parameter deviations are neglected. However, sensors, technical advances and the combination of production parameters and quality measures enable data-driven sampling strategies \cite*{MILO2015159}. 

The increasing amount of collected and interconnected production data and recent advances in machine learning enable data-driven models for quality inspection and may offer a potential solution for a smart sampling strategy. The models can include information about the produced pieces, changes in production parameters and increased parameter deviations. However, while those data-driven models may have great potential to improve the sampling strategy, their performance strongly relies on available training data, which is often limited. Therefore, we suggest a smart active sampling strategy for quality inspection to increase the model reliability over time and further enable the model to adapt to new production conditions. Smart active sampling is inspired by the field of active learning, where the learning algorithms query data points with a high prediction uncertainty to a user to create a ground truth and retrain themselves with this new information. Adapted to quality control in production lines, the machine learning model aims to detect low-quality units. The machine learning model input can be for example a set of measured production parameters. If the model has learned those production parameters well, it can detect low-quality parts with a high degree of certainty. However, if a novel combination of production parameters arrives, the model cannot detect it with high certainty and therefore query a part for quality testing. The smart active sampling strategy therefore only samples those parts for quality testing that are likely of low quality or where the quality cannot be predicted. This 1) reduces the number of samples and 2) allows continuous improvement of the collected samples and adaptation of the sampling strategy to new production conditions.

\section{Background}

Quality inspections aim to identify lots that do not meet quality standards as fast and reliable as possible while remaining at a low-cost level. Two major quality inspection tests are employed to achieve this goal: In-line quality inspection and out-line quality inspection. \textbf{In-line quality inspection} is a form of nondestructive examination and can be tested in-line on specific criteria such as optical techniques, including computer vision, spectral imaging, or near-infrared technology \cite{ZHANG2018213} and can be applied to every part. \textbf{Out-line quality inspection}  is based on a predefined sampling strategy that decides which parts are checked. Advanced in-line quality inspections help notice quality deviations, leading to faster feedback loops and therefore reduce the number of defective parts produced. However, quality criteria such as destructive or metallurgic tests that cannot be evaluated in-line still present challenges for controlling and maintaining production assets. However, these often necessary out-line inspections cause a temporal delay of inspection results, which may cause prolonged low quality or scrap production. 

\subsection{From Manual Inspections to Smart Sampling}
Modern digitalization and automatization led to a vast increase in process data obtained during production. The increase resulted in data-enabled data-driven models for condition monitoring \cite{wang2006condition,randall2021vibration} and anomaly detection \cite{LINDEMANN2019313,THEISSLER2017163}. 
Therefore smart sampling-based quality control is becoming increasingly important for quality inspections in production. In production lines and manufacturing processes, this is often done by monitoring relevant system parameters. The goal is to detect system or process deviations from normal behavior \cite{MILO2015159}. 

\subsection{Statistical Sampling Approaches}
Statistical sampling approaches often define normal measurements as statistical distributions for a given system parameter and then detect deviations from normality-testing a hypothesis. One of the first methods that used small samples to derive information on the production was Statistical Process Control (SPC) and is now widespread in the automotive industry \cite{godina2016quality,MILO2015159}. Statistical and machine learning based models that monitor production parameters and provide instant assessments can reduce the response time to defects due to early detection.

\subsection{Machine Learning Model based Sampling Approaches}
A typical scenario for machine learning models in production is to predict the outcome of a potential product quality inspection. Such an approach uses data obtained during production to estimate whether a part passes quality inspections. Supervised machine learning approaches require annotated training data to learn the system's actual behavior. Typically the outcome of quality inspections is only available for a subset of the produced parts. Once trained, the benefit of such a system is that an inline quality assessment is available for 100 $\%$ of the produced parts with little to no marginal costs. However, this traditional machine learning approach faces some difficulties, as discussed in the following.

\section{Current Challenges in Smart Sampling}\label{challenges}
While both statistical and machine learning-based methods show great potential in smart sampling strategies in quality inspection, there are some challenges to address. Therefore, we discuss the obstacles for implementing such a system in a real-life production environment from a machine learning perspective. 
As our working hypothesis for a smart sampling scenario, we assume that data from production and quality inspections are available and can be linked and that some causal relationships can be exploited by machine learning.   Further, we assume that the number of defective articles and the number of articles that can be sent to quality inspection is low compared to the overall number of produced parts.

\subsubsection{Price of Sampling due to destructive testing}

The cost of destructive sampling is comprised of the sampling period and destructive unit costs, including working hours and material. A large sampling period reduces the sampling costs; however, it increases the chance that defective items are delivered or further processed and consequently yields higher costs later. In the literature, this is also referred to as costs of false negatives \footnote{False negative: type I—error of rejecting the null hypothesis when it should be accepted.}. Costs that occur if insufficient quality parts are not detected. Creating a periodic sampling plan is therefore considered a complex optimization problem, in which the inclusion of active learning potentially enables an entirely new optimum.

\subsubsection{High variance in production parameter due to different products.}
Production lines are getting increasingly more complex, they have to meet new requirements and produce different products and variations. With every change in production parameters, new errors can occur and an active sampling strategy may be a good response to this dynamic environment.

\subsubsection{Complex Production Procedures and Increasing Number of Machine Learning Parameters}
In machine learning, the ratio between model complexity and available training data is ideally as small as possible. The model complexity, however, is increased with the number of process parameters included. Reducing or adapting the process parameters over time requires a more dynamic machine learning method, which is able to adapt to changes in the production environment actively.

\subsubsection{Missing or Insufficient Ground Truth Data}
Most models require significant annotated training data from actual out-line quality inspections. Therefore it requires a significant amount of time to collect enough data to train models with the required accuracy. Therefore, we require a new smart sampling method that addresses this issue by selecting the most informative data for training. 

\subsubsection{Data Imbalance: Class Imbalance and Data Distribution}
Typically, a data set for the prediction of quality inspection results will have a small proportion of data from defective parts and therefore be difficult to model. If we assume that the occurrence of defects and the sampling of parts for quality inspections are independent and random, the conditional probability of a defective part being randomly selected for inspection becomes very low. Therefore, we can expect to improve the accuracy of the machine learning model by providing more samples of defective parts \cite{DBLP:journals/corr/abs-1305-1707}.

\subsubsection{Model Centric-AI}
Current statistical and machine learning approaches are often model centered, focusing on the right model approach rather than on data itself. However, it is essential to ensure data quality improvement in such approaches. A smart active learning approach may enable such a focus shift to a more data-centric approach \cite{9671795}.

\section{Smart Active Sampling}
Based on the challenges in smart sampling discussed in the previous section \ref{challenges}, we propose a new sampling strategy for quality testing, which allows for more efficient testing while also increasing model reliability. Our new sampling strategy is inspired by the field of active learning. 
First, we introduce our approach, followed by the description of adapting smart sampling to quality control.

\begin{figure}[ht]
    \centering
    \includegraphics[width=0.95\columnwidth]{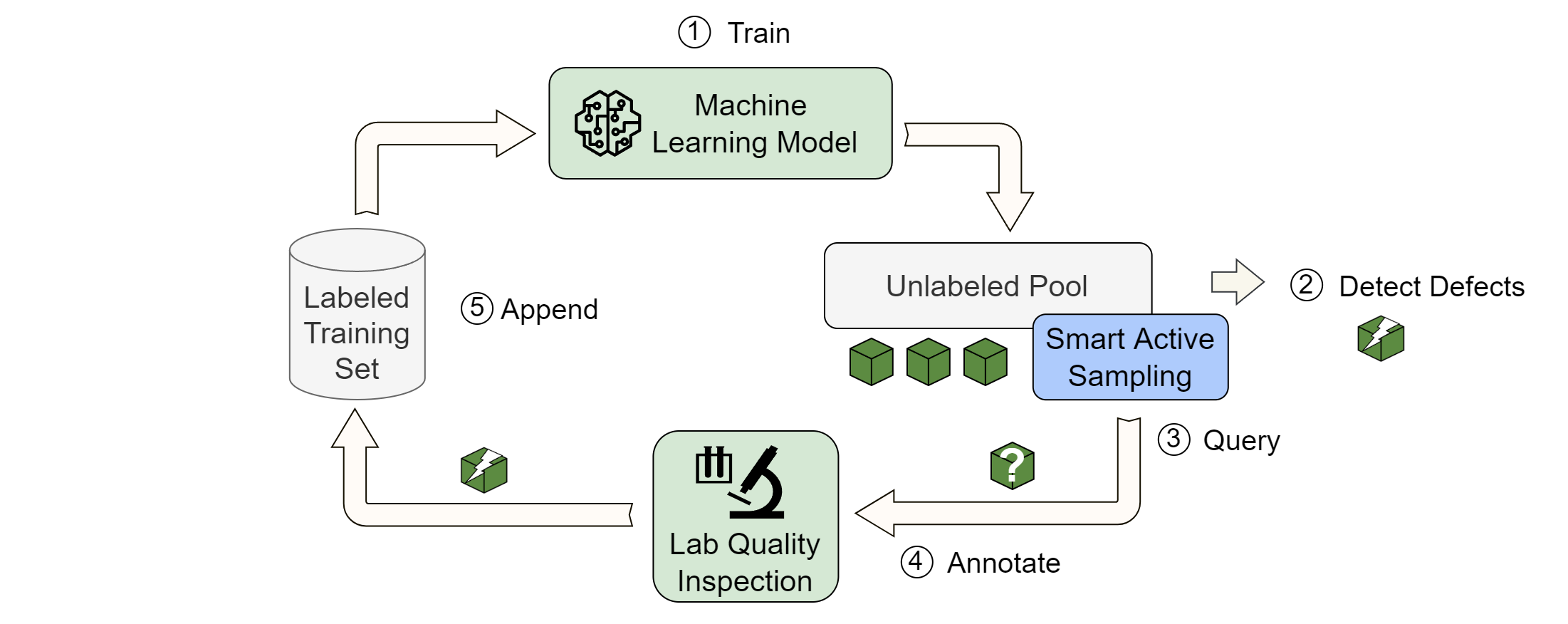}
    \caption{The Smart Active Sampling loop. (1) Starting from a pre-trained machine learning model. (2) Applying the machine learning model to an unlabeled pool of samples and detecting defect parts. (3) Select those samples with a low model confidence and send them to lab inspection. (4) Annotate data at the lab by experts. (5) Append the newly labeled sample to the labeled training data set. }
    \label{fig:my_label}
\end{figure}

Machine learning models work better when the number of annotated samples is higher. In many scenarios, annotations come with non-neglective costs, and it is impossible to label all samples. Thus, a labeling strategy to select which samples to annotate is required. Random sampling is a simple and effective strategy typically used in industrial quality inspections but can be less practical for unbalanced data.

Active learning is a particular case of machine learning in which a learning algorithm can select when to query for a new label. By choosing the most useful samples for annotation, the number of samples for an effective machine learning model can be much smaller than with random sampling. The definition of useful depends on the implementation and is selected for a specific purpose. Typical goals are minimizing the generalization error or reducing the uncertainty of the model. \cite{settles2012active,settles2009active}

We believe that active learning is an excellent approach for improving the effectiveness of a machine learning approach in quality assurance by letting the model decide which data points to annotate. We also expect several advantageous side effects, which we elaborate on in the following sections. We refer to our variant of active learning adapted to sample selection for quality inspection as \textbf{\textit{Smart Active Sampling}}.

\subsection{From Active Learning to Smart Active Sampling}

Our machine learning model aims to detect low-quality units by adapting the smart active learning approach to quality control in production lines.
The machine learning model takes the available input parameters from the production process and predicts the quality measures. Input parameters can be measurements of physical processes in the production line, such as injection pressures, adjustment values of the machine, information about the parts to be processed, and many more. Using active learning a model can be trained on the available input parameters which can then predict the quality with a high degree of certainty for almost all produced parts. If a combination of production parameters arrives, where the model predicts a faulty part or it cannot predict the product quality with high certainty, the learning model sends the part to quality inspection. 
This smart active sampling approach tackles the problems mentioned in the previous section and can be summarized as follows.

\subsection{Benefits of Smart Active Sampling}
\input{figures/tikz1}
\subsubsection{Efficient Collection of Ground Truth Data}
Smart active sampling leads to a more efficient collection of well-balanced ground truth data. The learning model only sends items with unfamiliar production parameters to quality inspection, where it can not predict the quality measures with high enough confidence. Unnecessary inspections for parts with well-known parameters and settings are avoided; therefore, duplicated entries in the ground truth data are minimized.

\subsubsection{Reduced Data Imbalance}
\begin{figure}
    \centering
    \includegraphics[width=0.8\columnwidth]{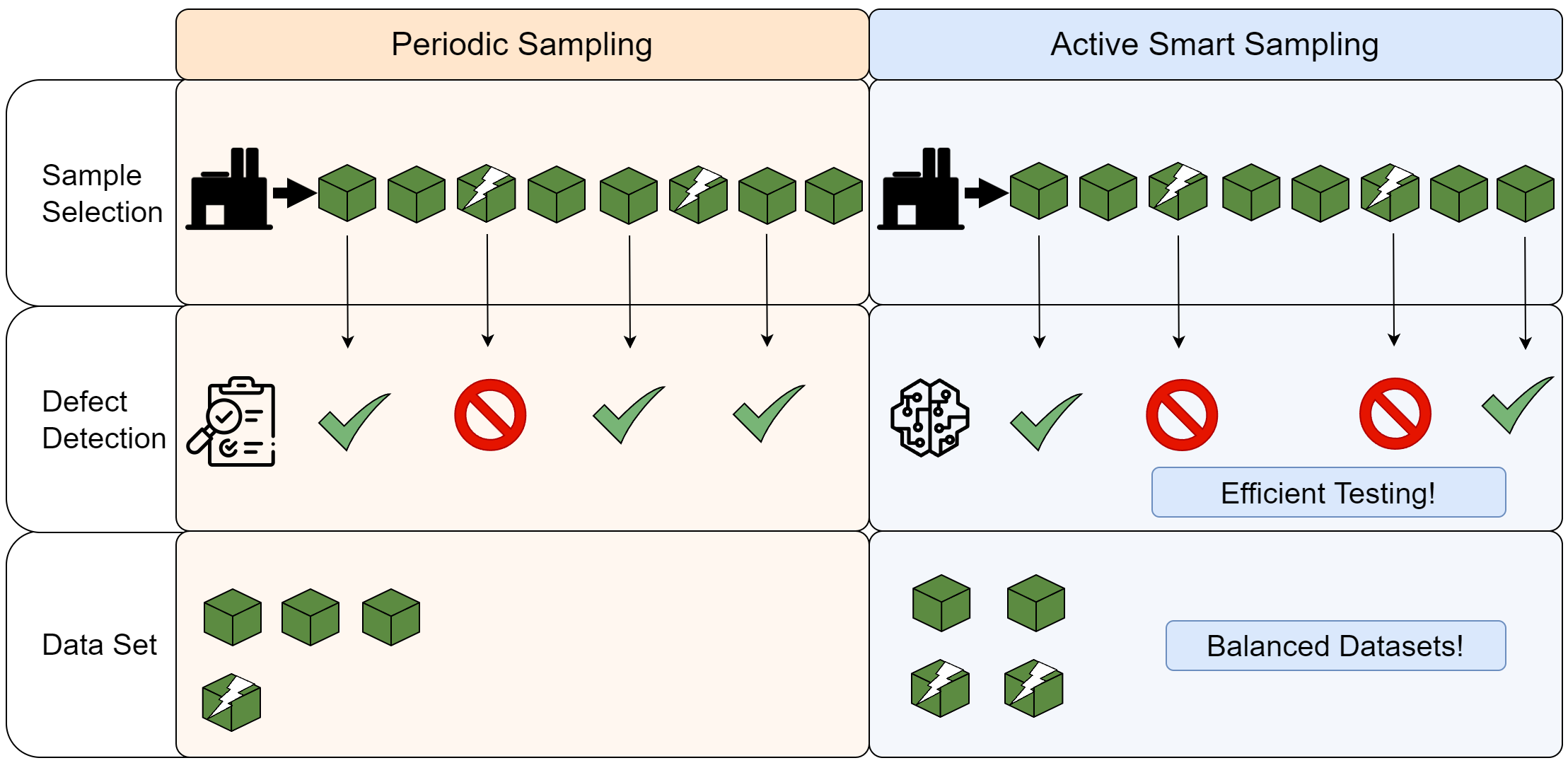}
    \caption{Smart Active Sampling allows for more efficient testing. The collected data is well balanced and the values of the input parameters are evenly distributed, leading to more accurate and more robust models. }
    \label{fig:imbalance}
\end{figure}

A major advantage of the smart active sampling approach is the increase in data quality. Active learning by construction leads to a more balanced data set where the target categories OK vs NOT OK are more evenly distributed, as depicted in Fig. \ref{fig:imbalance}. The same holds true for the input parameters, which are also more evenly distributed. Consequently, better models can be trained with less training data. 

\subsubsection{Data-Centric AI}
Our smart active sampling proposal resembles the philosophy of the data-centric AI approach advocated by Andrew Ng for industrial applications where only a limited amount of training data is available\footnote{A good summary can be found at https://landing.ai/data-centric-ai/.}. Data-centric AI focuses on data instead of code, where the data conveys all the information the model must learn. The data used to build AI systems is systematically engineered, leading to more robust models with marginal sensitivity to data drifts and variances in production parameters. 
With this approach, better models can be built with less data \cite{datacentricNN}, leading to new applications in areas where previously not enough data was available or too expensive to obtain.

\subsubsection{Time to Deployment}
Time to deployment can be a major hurdle for implementing machine learning systems in production environments. The aforementioned improvements also reduce the time to deployment and productive use of the system. Efficient collection of ground truth data that is well balanced allows for building more accurate models significantly faster than before.

\subsubsection{Reduced Price of sampling}
Building better models with less training data results in less destructive testing until the deployment of the model. The number of suitable, balanced training data increases the performance and certainty of the machine learning model over time. The model can make better and better predictions and the number of inspections and even disruptive tests may be reduced over time. Once the model is deployed, smart active sampling has two advantages over previous sampling methods.

\textbf{Efficient testing}: If the production runs smoothly without producing scrap parts, smart selection for destructive testing ovoids unnecessary quality inspections and minimizes inspection costs. 

\begin{figure}[ht]
    \centering
   \includegraphics[width=1.0\columnwidth]{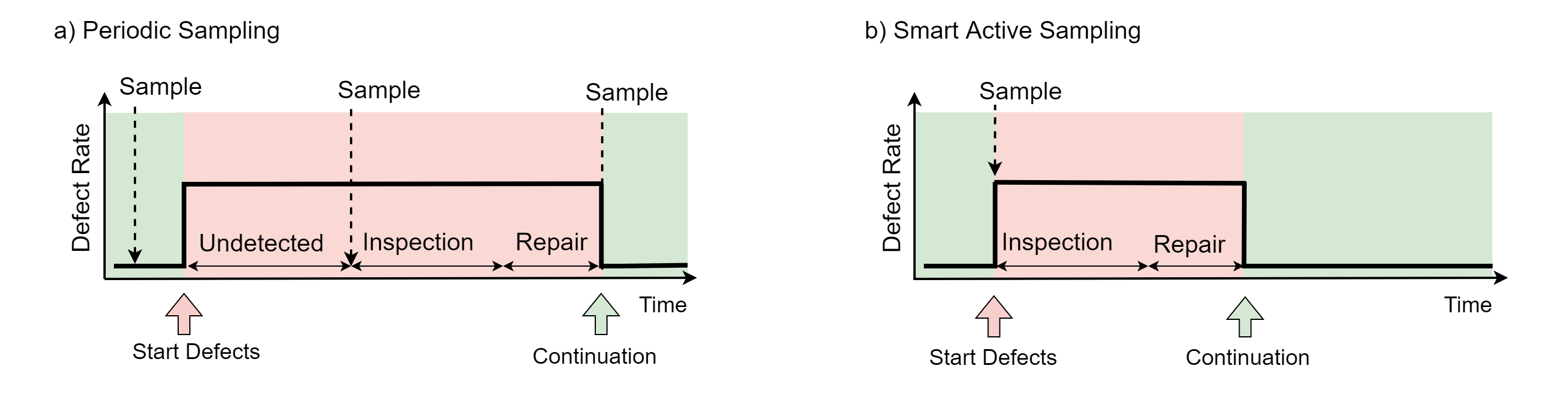}
    \caption{Time to response: With smart active sampling undetected production of scrap parts is minimized because defective parts are detected in time by the system and sent to quality inspection.}
    \label{fig:response}
\end{figure}

\textbf{Time to react}: If the model detects inconsistencies in the production and is unsure of the quality, parts are sent to quality inspection immediately. Scrap parts or a drift towards quality violations are detected and the appropriate actions, avoiding further production of scrap parts, can be taken (see fig. \ref{fig:response}). This is in stark contrast to fixed interval sampling, where in the worst-case scenario, as many scrap parts as the interval length are produced. Early detection of problems also minimizes machines' downtime, often resulting from the production of scrap parts.

\subsection{Practical Recommendation for Implementation}

Finally, we give a few practical recommendations to accelerate the implementation of the smart active sampling strategy.

Start as early as possible. In the best case scenario, start creating your models and the corresponding feature engineering already during the development process of the product, where it is crucial to involve the subject matter experts (SME) in the process.   
This increases the chance to get well-balanced data sets because during the development of the product and the running up phase of the production different calibrations and their influence on the quality are tested. For example, scrap parts can be produced on purpose to increase the input parameter range for the model. 
In addition, close collaboration between the SMEs and the data scientist to extract the most significant features helps to build more accurate models and may unveil new dependencies and correlations not known before. Careful feature engineering also supports the data-centric approach and makes the machine learning system more resilient to changes in the data. 

In contrast, when a machine has been running for years and the Overall Equipment Effectiveness is already on the upper end of the spectrum and the variance in the data is small and it would take a long time to collect enough data to build a model accurate enough for deployment. At this stage, it is also hard to involve SMEs in the model building due to their availability. Usually, SMEs do not have the capacity to work on well-running machines.

\section{Conclusion}
In this work, we proposed a new sampling strategy for out-line testing in quality inspection tasks based on active learning principles, which we call smart active sampling. The learning algorithm decides which parts are sent to quality inspections. Detected scrap parts are sent to quality inspection as well as parts where the algorithms are not sure, leading to a well-balanced and more evenly distributed data set. Consequently, more accurate and robust machine learning models can be built and deployed. 
Compared to random or fixed interval sampling, scrap parts or drifts towards quality violations are detected immediately. With this information, appropriate actions can be taken in time, resulting in higher availability of the machine and a better quality portion. Both availability and quality are components of the Overall Equipment Effectiveness (OEE); therefore, smart active sampling increases the OEE.
On the contrary, if no scrap parts are produced, no parts are sent to unnecessary expensive testing reducing the overall quality assurance costs.

\section{Acknowledgements}
This work was partly funded by the Austrian Research Promotion Agency (FFG) through the project INTERACTIVE (Project ID: 874042). We especially want to thank Stefan Kemptner for fruitful discussions and for providing insights into the challenges of quality assurance from an operational perspective.


  \bibliography{sample-base.bib}

\end{document}